%% file: main.tex
\pdfoutput=1

\documentclass[11pt, dvipsnames, table]{article}

\usepackage{acl}

\usepackage{times}
\usepackage{latexsym}
\usepackage[T1]{fontenc}
\usepackage[utf8]{inputenc}
\usepackage{microtype}
\usepackage{inconsolata}
\usepackage{graphicx}
\usepackage{amsmath}
\usepackage{bbm}
\usepackage{amssymb}
\usepackage{amsfonts}
\usepackage{booktabs}
\usepackage{lipsum}
\usepackage{multicol}
\usepackage{makecell}
\usepackage{rotating}
\usepackage{multirow}
\usepackage{utils}
\usepackage{caption}
\usepackage{subcaption}
\usepackage{xcolor}
\usepackage{enumitem}
\usepackage{linguex}
\usepackage{todonotes}
\usepackage[normalem]{ulem}
\usepackage{fdsymbol}
\usepackage{wasysym}

\definecolor{blu}{HTML}{3D85C6}
\definecolor{yell}{HTML}{F1C232}
\definecolor{orang}{HTML}{d95f02}
\definecolor{purpl}{HTML}{7570b3}

\makeatletter
\newcommand*\iftodonotes{\if@todonotes@disabled\expandafter\@secondoftwo\else\expandafter\@firstoftwo\fi}  
\makeatother

\newcommand{\comps}{\textsc{comps}}
\newcommand{\compsqa}{\textsc{comps-qa}}
\newcommand{\first}{\textcolor{blu}{\textsc{first-correct}}}
\newcommand{\recent}{\textcolor{yell}{\textsc{recent-correct}}}
\newcommand{\compswugsdist}{\textsc{comps-wugs-dist}}

\title{Experimental Contexts \emph{Can} Facilitate Robust Semantic Property Inference in Language Models, but Inconsistently}

\author{Kanishka Misra$^\tau$ \quad Allyson Ettinger$^\alpha$ \quad Kyle Mahowald$^\tau$\\
    $^\tau$The University of Texas at Austin \quad $^\alpha$Allen Institute for Artificial Intelligence\\
    \texttt{kmisra@utexas.edu} \quad \texttt{allysone@allenai.org} \quad \texttt{kyle@utexas.edu}}

\begin{document}
\maketitle

\begin{abstract}
Recent zero-shot evaluations have highlighted important limitations in the abilities of language models (LMs) to perform meaning extraction.
However, it is now well known that LMs can demonstrate radical improvements in the presence of experimental contexts such as in-context examples and instructions. 
How well does this translate to previously studied meaning-sensitive tasks?
We present a case-study on the extent to which experimental contexts can improve LMs' robustness in performing property inheritance---predicting semantic properties of novel concepts, a task that they have been previously shown to fail on. 
Upon carefully controlling the nature of the in-context examples and the instructions, our work reveals that they can indeed lead to non-trivial property inheritance behavior in LMs. 
However, this ability is inconsistent: with a minimal reformulation of the task, some LMs were found to pick up on shallow, non-semantic heuristics from their inputs, suggesting that the computational principles of semantic property inference are yet to be mastered by LMs.
\end{abstract}

\section{Introduction}

Carefully controlled behavioral analyses on meaning-sensitive tasks have revealed holes in the ability of language models (LMs) to demonstrate robust meaning extraction and use \citep[][\textit{i.a}]{pandia-ettinger-2021-sorting, elazar-etal-2021-back, schuster-linzen-2022-sentence, weissweiler-etal-2022-better, misra-etal-2023-comps, kim-schuster-2023-entity}.
However, since a large subset of these investigations uses zero-shot evaluation as the primary methodology,
there are growing concerns that they do not paint a complete picture of LMs' abilities \citep{lampinen2022recursive,sinha-etal-2023-language}.
Conclusions that LMs lack a particular ability may be overhasty if it turns out the ability is easily accessed through in-context learning, different question formulations, or particular instructions \citep{lampinen2022recursive, wei2022chain}.

\begin{figure}[!t]
    \centering
    \includegraphics[width=\columnwidth]{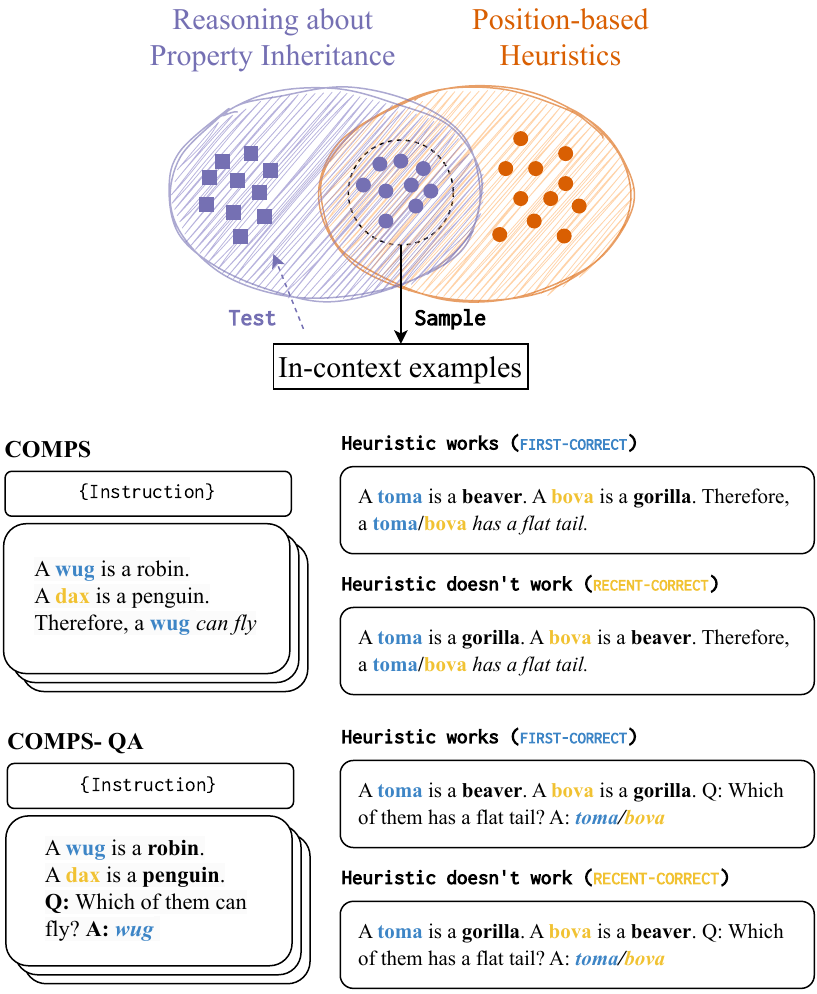}
    \caption{
    LMs are prompted with in-context examples that are compatible with both, \textcolor{purpl}{robust property inheritance}, as well as \textcolor{orang}{position-based heuristics}. At test time, we evaluate on cases where the heuristics support desirable behavior \textit{and} on cases where they do not. We use stimuli from \comps{} and its reformulation as a QA task. 
    }
    \label{fig:paradigm}
    \vspace{-1.2em}
\end{figure}

Our focus\footnote{Our code can be found \href{https://github.com/kanishkamisra/fewshotcomps}{in this repository}} in this paper is a particularly challenging dataset for meaning-sensitive behavior: \comps{} \citep{misra-etal-2023-comps}, which contains minimal pair sentences that test the ability of LMs on property knowledge of everyday concepts (\textit{a \textbf{beaver}/gorilla has a flat tail}) and their inheritance for novel concepts (\textit{a wug is a \textbf{beaver}/gorilla.~therefore, a wug has a flat tail}). 
Contemporary LMs failed miserably on the hardest subset of the \comps{} stimuli, the examples of which contain two novel concepts (\textsc{wug} vs. \textsc{dax}), where only one of them inherits the target property (\textit{has a flat tail}): 

\ex.\label{ex:compswugs} A \textcolor{blu}{\textbf{wug}} is a beaver. A \textcolor{yell}{\textbf{dax}} is a gorilla. Therefore, a \textbf{\textcolor{blu}{wug}/\textcolor{yell}{dax}} has a flat tail.

Given the success of LMs on a wide variety of complicated tasks, their utter failure on this seemingly straightforward task remains puzzling.
Here, we systematically explore \comps{} on 12 LMs ranging from 1.5--13B parameters, varying (a) whether they are evaluated zero-shot or with multiple examples, and (b) whether or not instructions are present.

Unlike other minimal-pair datasets, 
using \comps{} in an in-context learning setting is non-trivial (and thus potentially informative). 
This is because the task can be solved using a position-based heuristic.
For example, in one subset of \comps{}, the target property is always attached to 
\textbf{first} novel concept---like in \cref{ex:compswugs}. 
Importantly, LMs' failures on \comps{} were shown to be in part a result of models' tendencies towards heuristic behavior: the performance of autoregressive LMs was found to be particularly bad when the distractor (\textit{a dax is a gorilla}) is \textit{recent}---i.e., they showed a recency bias in attributing properties to novel concepts.
In that sense, \comps{}  follows a rich body of work in which tasks are set up in a manner that two types of generalization mechanisms can lead to the same prediction, but only one of which is desirable \citep{mccoy-etal-2019-right, mccoy2020does, warstadt-etal-2020-learning, mueller-etal-2022-coloring, si-etal-2023-measuring}.

We find that experimental contexts, as operationalized using in-context examples and instructions, can in fact demonstrate robust improvements in LMs' property inheritance behavior as measured by \comps{}. 
However, this improvement comes with a caveat: 
With a minimal reformulation of \comps{} into a QA task, where there is a direct link between the LMs' output space and the features of the input that control the heuristic, many LMs showed a strong preference towards the heuristic, and were therefore at chance.
Interestingly, LMs that receive explicit supervision to follow instructions do show some resistance to positional heuristics, but not always---occasionally, certain instruction-tuned LMs also tend to show a preference for one particular type of heuristic. 
These discrepancies in LMs' performance underscore their difficulty of mastering the reasoning ability to robustly demonstrate semantic property inheritance.

\section{Methodology}
\paragraph{Dataset}
We use the most difficult subset of the \comps{} dataset \citep{misra-etal-2023-comps}---\compswugsdist{}. This dataset contains 13,828 sentence pairs of the form similar to \cref{ex:compswugs}, constructed using 152 animal concepts and 991 properties.

\paragraph{Stimuli re-design}
We take a number of steps to minimize noise from other (likely uninterpretable) heuristics beyond the ones we have set out to target.
First, we enforce that the concepts and properties that appear in the in-context examples are disjoint from ones that are used in tests. To this end, we sample 50 concepts and their relevant properties and reserve it for our in-context examples, leaving the rest to be sampled for our test set. 
We also enforce this constraint for our novel concepts---i.e., all in-context examples contain different nonce words, and the collection of nonce words for the in-context examples and the test set is disjoint.
Furthermore, we counterbalance the nonce words in the test set in a manner that having a bias towards one of them would lead to chance performance.
We additionally also use multiple different sets of in-context examples, to add variability and to ensure that the results are not only due to one particular choice of in-context examples. 
In total, we use 10 different in-context learning example sets, each containing 6 different \comps{} stimuli. 
For our test set, we use a constant set of 256 unique pairs sampled from our pool of stimuli containing unused concepts and properties.

\paragraph{Heuristics} Our most important design decision is to consider two distinct sets of stimuli---each separately making available the two types of heuristics that the LMs could rely upon: \first{} and \recent{}, where the property is inherited by the \textcolor{blu}{first} and the \textcolor{yell}{most recent} novel concept, respectively. 
That is, for the same set of in-context examples, we have a version where the \textcolor{blu}{first concept} is correct like in \cref{ex:compswugs}, and one where the \textcolor{yell}{most recent concept} is correct:

\ex.\label{ex:compswugsrecent} A \textcolor{blu}{\textbf{wug}} is a gorilla. A \textcolor{yell}{\textbf{dax}} is a beaver. Therefore, a \textbf{\textcolor{blu}{wug}/\textcolor{yell}{dax}} has a flat tail.

For each type of in-context stimuli, we similarly have two versions of test stimuli: one that is consistent with the target heuristic, and one that is not. 
That is, a test example that is consistent with the \first{} heuristic will also have its \textcolor{blu}{first concept} be the one that inherits the property in question, while one which is inconsistent will have the \textcolor{yell}{most recent concept} be the inheritor of the property.
Therefore, a model that shows a preference for a given heuristic will succeed only on one test set and succumb on the other, while a model that is robust to the heuristics will succeed on both.

\paragraph{Reformulation into QA} 
The original \comps{} stimuli test for property inheritance using declarative statements, where models are tested for the log-probability they asign to the property (\textit{has a flat tail}) given either of the two concepts (\textit{wug} vs. \textit{dax}). 
Here we additionally consider an alternate formulation of \comps{} as a question answering task (\compsqa{}), where we make the property explicit in the prompt to the model and instead ask which of the two concepts possesses it:

\ex.\label{ex:compsqa} A \textcolor{blu}{\textbf{wug}} is a beaver. A \textcolor{yell}{\textbf{dax}} is a gorilla. \textbf{Question:} Which one of them has a flat tail? \textbf{Answer:} \textbf{\textcolor{blu}{wug}/\textcolor{yell}{dax}}

Since the shallow heuristics we consider are controlled by the relative ordering of the novel concepts, this formulation of the task directly allows us to link the models' output space (the novel concepts) to the heuristics (positions).

\paragraph{Testing setup}
For the original \comps{} setting we follow \citet{misra-etal-2023-comps} and compare the log-probability of the property phrase given the correct vs. the incorrect prefix. 
For \compsqa{} however, since we have a constant prefix (same premises and question), we evaluate the relative log-probability of the two novel concepts, only one of which is the correct answer. 
Accuracy in both cases is the proportion of cases the correct surface form was assigned relatively higher log-probability. 
Since we use pairwise comparisons throughout, chance performance is 50\%.

\paragraph{Instructions}
We consider four different kinds of instruction templates, with varying levels of detail (see \cref{sec:instructions}) per formulation (\comps{} and \compsqa{}). In our experiments we report results on the instruction that gives the best average performance for a given model.

\paragraph{LMs tested}
We evaluated 8 different open-source LMs, all of which are decoder-only, and were accessed using the huggingface hub \citep{wolf-etal-2020-transformers}: \textbf{GPT-2 XL} \citep{radford2019language}; \textbf{OPT-6.7b} \citep{zhang2022opt}; \textbf{Llama-2} \citep[we used the \textbf{7B} and the \textbf{13B} versions;][]{touvron2023llama}; \textbf{Mistral-7B} \citep{jiang2023mistral}; \textbf{Llama-3-8B} \citep{dubey2024llama}; \textbf{OLMo-7B} \citep{groeneveld2024olmo}; and \textbf{Gemma-2-9B} \citep{team2024gemma}. In addition to these 8, we tested the instruct-tuned versions of OLMo, Llama-3-8B, Mistral-7B (v0.3), and Gemma-2-9B, to analyze if instruction tuning affects LMs' robustness. Details about all 12 LMs can be found in \Cref{sec:implementation}.

\begin{figure*}[!ht]
    \centering
    \includegraphics[height=8.2cm]{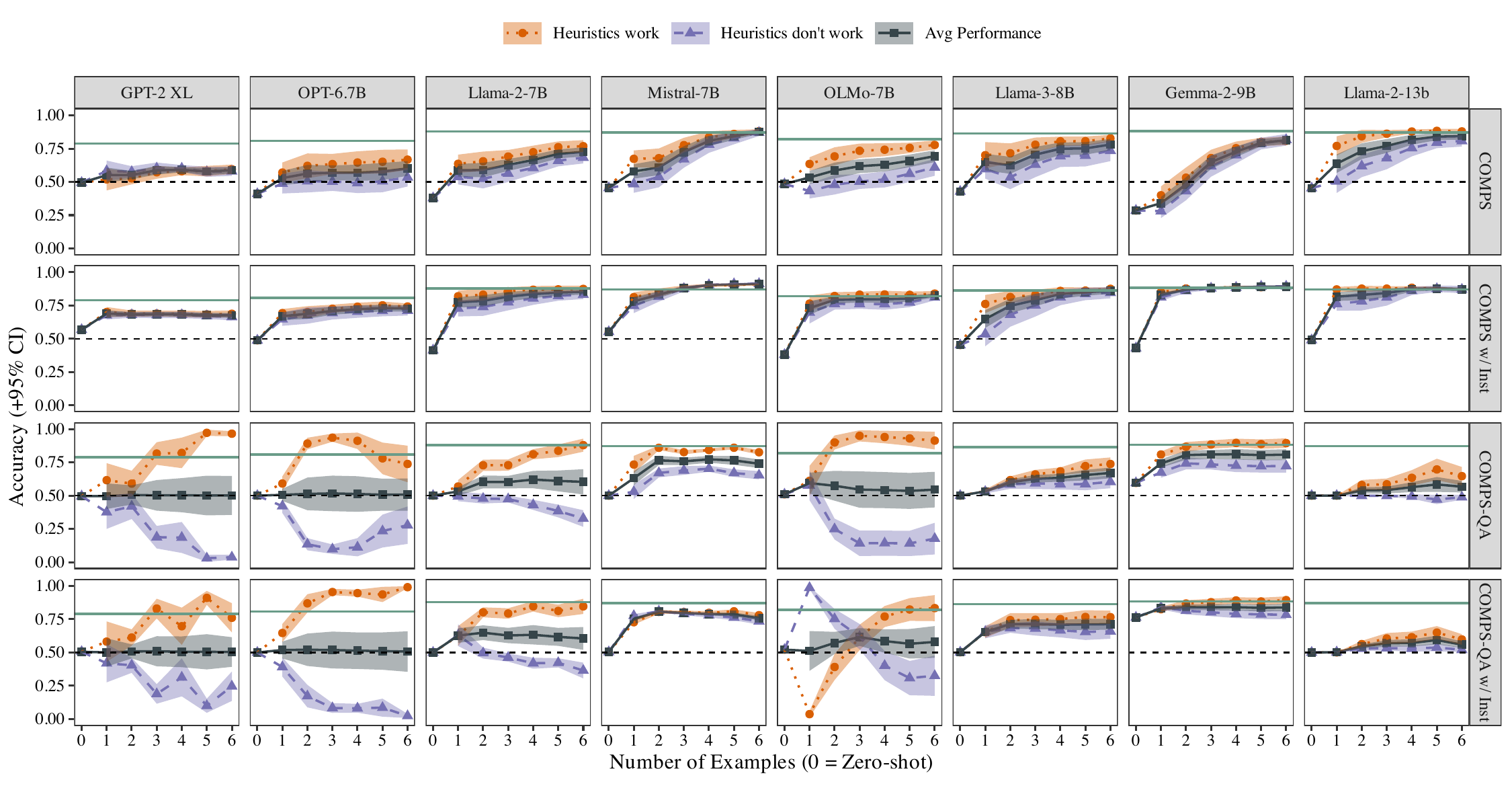}
    \caption{Overall results from our experiments testing non-instruction tuned LMs on \comps{} and \compsqa{} using in-context examples, with and without instructions. Results are aggregated across both heuristics: \first{} and \recent{}. Error bars are over different sets of in-context examples. Most models start off near chance in the 0-shot case, but many improve as more examples are given. 
    Solid green line depicts each model's base property knowledge performance, while the black dashed line depicts chance performance.}
    \label{fig:main-results}
    \vspace{-1em}
\end{figure*}

\section{Analyses and Results}

We evaluate on \comps{} and \compsqa{}, with and without instructions. In each case, we progressively supply 0 through 6 in-context examples, allowing us to track the dynamics of the models' performance with an increasing amount of demonstrations. 
Together with our separate types of test sets and heuristics encoded in the in-context examples, along with five different instruction settings (four with and one without) we run 2420 experiments per LM.
We hypothesize that LMs would be more sensitive to the positional heuristics in \compsqa{} because of the clear link between their output space and the relative position of the novel concepts---the feature that controls our target heuristics.

\begin{figure}
    \centering
    \includegraphics[width=\columnwidth]{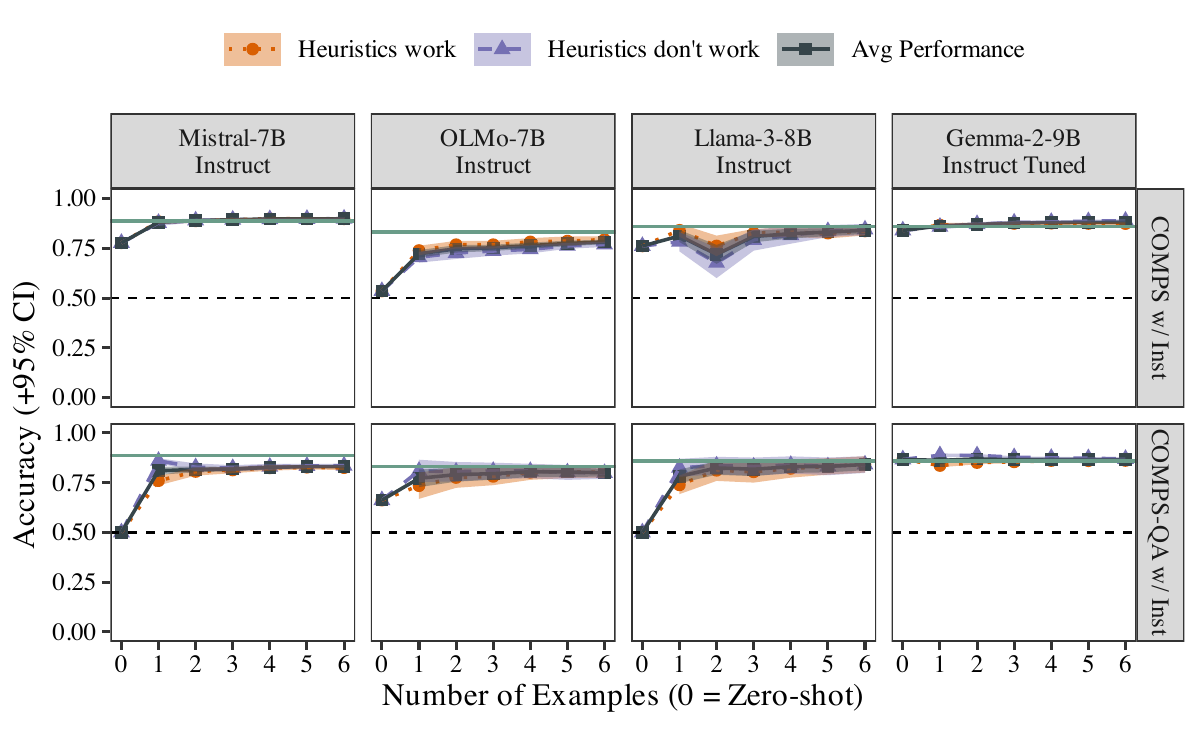}
    \caption{Overall results on the four instruction-tuned models considered. Results are aggregated across both heuristics: \first{} and \recent{}. Error bars are over different sets of in-context examples. Solid green line depicts each model's base property knowledge performance, while the black dashed line depicts chance performance.}
    \label{fig:instruct-tuned}
    \vspace{-1em}
\end{figure}

\Cref{fig:main-results} shows accuracies of the tested LMs on our four different \comps{} settings as a function of the number of in-context examples provided to them, for both: cases where the \textcolor{orang}{\textbf{heuristics are consistent with success on the test set}}, and cases where \textcolor{purpl}{\textbf{they are not}}. 
We also show an additional curve denoting the average performance across the two types of test sets to paint an overall picture of the models' performance. 
In this figure, the extent to which a model relies on heuristic is indicated by the gap between the \textcolor{orang}{dotted ($\newmoon$)} and the \textcolor{purpl}{dashed ($\blacktriangle$)} lines. A model that is robust to the heuristics will have curves of both colors rise above chance, with no gap between the two, while one that is prone to using heuristics will have its \textcolor{orang}{dotted ($\newmoon$)} curve be substantially greater than its \textcolor{purpl}{dashed ($\blacktriangle$)} curve. Following the same format, \Cref{fig:instruct-tuned} shows results on the four instruction-tuned models evaluated. In all these plots, we also visualize the LMs' `base' property knowledge, where we compare (in a zero-shot manner) the LMs' preference for attributing the properties in the test set to the correct `known' concepts (i.e., beaver vs. gorilla in our examples).

\paragraph{Experimental context can improve attribution of properties to concepts...}
{On \comps{}, models unsurprisingly start off at chance performance on average, corroborating the previous findings of \citet{misra-etal-2023-comps}. 
However, in the presence of in-context examples and instructions, they are able to improve monotonically as the number of in-context examples increases, reaching similar levels of performance on property inheritance as that of the models' base knowledge. 
It is worth noting \textbf{Llama-2-13b} and \textbf{OLMo-7B} do occasionally show a preference for heuristics in the absence of instructions.
An intermediate conclusion that we draw here is that LMs can indeed demonstrate non-trivial property inheritance on observing a few examples that reflect that behavior.}

\paragraph{...but not the attribution of concepts to properties}
While experimental context seems to aid models in attributing properties to the right concept in context, the same does not hold on \compsqa{}. 
Similar to \comps{}, models start off at chance performance on average with a zero-shot set up, however, unlike in the case of \comps{}, many LMs seem to consistently prefer the heuristics available in the prompt, showing, at times, worse than chance performance on cases where the test set does not follow the heuristic.
This is most apparent for \textbf{GPT-2 XL}, \textbf{OPT6.7b}, \textbf{Llama-2-7b}, \textbf{OLMo-7B}---here there is a substantial gap between the accuracy for cases where heuristics support performance on the test set and the accuracy for cases where they do not. Other models, like \textbf{Mistral-7B} and \textbf{Gemma-2-9B} occasionally show behavior compatible with the heuristics, but this does not hurt their overall performance, which is consistently above chance, and close to their base knowledge.

Our results suggest that LMs are more likely to show behavior that is compatible with the \textit{use} of positional heuristics when their output space (choice between the two novel concepts) has a clear connection with positional artifacts in their input (relative ordering of the novel concepts). This is consistent with our hypothesis in about 11 out of 16 cases (where for 9 of the cases, heuristics end up drastically hurting performance). 
When this link is not clear and models must instead predict likely properties given a novel concept (i.e., in \comps{}), instructions and in-context examples do seem to lead to robust performance.
It is important to note that instructions \textit{alone} do not always account for the observed improvement---LMs' performance on zero-shot settings are consistently still at chance in all cases, suggesting that it is their interaction with in-context examples that critically alters models' output distribution to support desirable property inference behavior.

\paragraph{Benefits of instruction tuning...} Common to the previous analysis is that the tested LMs were all pretrained using the standard next-word prediction objective. While instruction-following capabilities have been shown to arise in such models, a new paradigm has emerged in the field that tunes such pre-trained LMs post-hoc to explicitly follow instructions using a plethora of different techniques not limited to those that leverage feedback from human-provided labels \citep{ouyang2022training}. How does instruction tuning interact with the presence of position-based heuristics? \Cref{fig:instruct-tuned} shows results on four LMs which are instruction-tuned versions of some of the LMs used in the previous analysis, for \comps{} and \compsqa{} (with instructions). We find these four LMs to show consistently above-chance performance on both datasets, suggesting non-trivial robustness brought about by instruction-tuning. Interestingly, while most models show improvements from zero-shot (often times at chance), the instruct-tuned \textbf{Gemma-2-9B} model shows no signs of improvement, instead showing nearly constant performance on both \comps{} and \compsqa{} regardless of the number of in-context examples.

\begin{figure}
    \centering
    \includegraphics[width=\columnwidth]{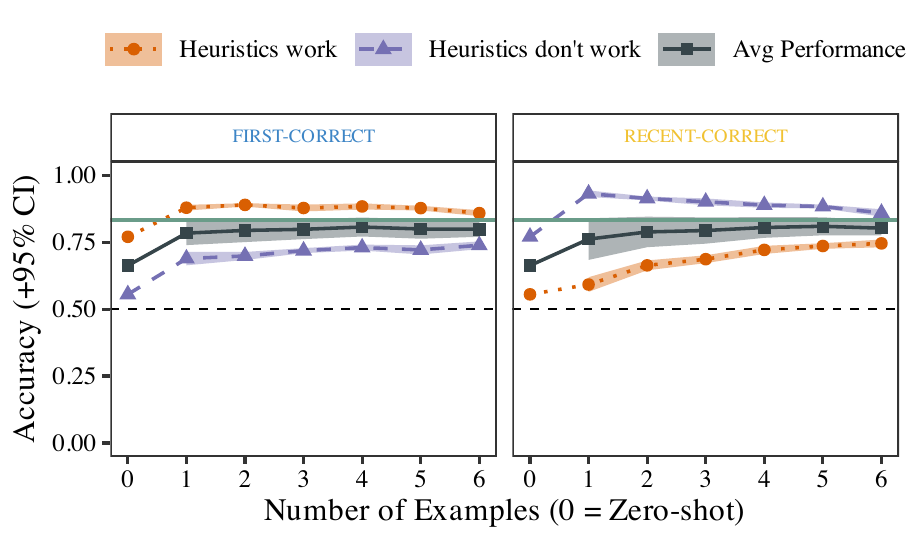}
    \caption{Finer-grained results on Instruct-tuned \textbf{OLMo-7B} LM demonstrating its preference for selecting the first concept, regardless of the heuristics.}
    \label{fig:olmo-special-case}
    \vspace{-1em}
\end{figure}

\paragraph{...and its occasional pitfalls} While \Cref{fig:instruct-tuned} shows that LMs benefit from instruction-tuning in the context of robust property inheritance, a detailed breakdown of the results offers a few caveats. In \Cref{fig:olmo-special-case} we show \textbf{OLMo-7B-Instruct}'s accuracies broken-down by heuristic on \compsqa{}, the setting where we hypothesized LMs to be most likely relying on heuristics. For the \first{} heuristic, the model shows a preference for the heuristic (greater performance when it supports success on the test set), while for \recent{}, we see the opposite to be true. Here, we observe that the LM does better when the heuristic does not support success on the test set, suggesting that the LM has a bias for preferring entities mentioned first---a bias so strong that it persists even when there is ample evidence in the context of the model that favors the opposite heuristic. This suggests that while instruction tuning leads to consistently above-chance performance on challenging property inheritance problems, it is not entirely robust to position-based heuristics.

\section{Conclusion}
We investigated the extent to which in-context examples and instructions---key components that drive impressive performance in contemporary LMs---can overcome important limitations of LMs at tests that have poked holes in their ability to extract conceptual meaning from text. 
As a case study, we analyzed how well such experimental contexts can improve LM abilities to perform property inheritance \citep{murphy2004big, misra-etal-2023-comps} in context: binding of novel concepts to existing concepts, and endowing them with valid property inferences as a result.

Our findings suggest that mastery of this ability has yet to be robustly achieved, and that many LMs trained on the next-word prediction objective are still prone to using shallower patterns in their context rather than systematically extracting conceptual meaning. There are several exceptions to this, however, for example, modern LMs such as Mistral-7B and Gemma-2-9b did show heuristic behavior, but were also well above chance, especially in the presence of instructions. Additionally, instruction tuned LMs were found to be substantially more robust than their non-instruction tuned counterparts, suggesting advantages to explicit instruction tuning. However, they too sometimes showed counterintuitive behavior, where their strong bias for a consistent position overcomes the heuristic patterns available in the input. We leave the nuanced exploration of this behavior induced by instruction-tuning for future work.

\section{Limitations}
\paragraph{Single dataset} A clear limitation of this work is that it exclusively focuses on a single dataset: \comps{} \citep{misra-etal-2023-comps}. So, a question that arises here is to what extent are our findings localized to the chosen dataset vs. meaning-sensitive evaluations in general. 
This would require a further non-trivial, non-straightforward amount of work, since: 1) different meaning sensitive evaluations focus on different (though equally useful) operationalizations of meaning; and more importantly 2) not all prior work in this area focuses on a standardized and well-defined usage of heuristics that is directly transferable to the experimental setup we have used in this work \citep[following][]{mccoy-etal-2019-right, mccoy2020does, warstadt-etal-2020-learning, mueller-etal-2022-coloring, si-etal-2023-measuring}. 
We do hope that our work contributes to the larger-scale vision of carefully benchmarking different types of meaning extraction abilities in LMs in a controlled manner.

\paragraph{Lack of mechanistic insight} Our work continues the long-standing precedent of using carefully constructed behavioral experiments to conclude about the competence of LMs \citep{linzen-etal-2016-assessing, gulordava-etal-2018-colorless, futrell-etal-2019-neural, ettinger-2020-bert, warstadt-etal-2020-blimp-benchmark}
However, recent works have made impressive strides in localizing the kinds of computations that give rise to the observed behavior in LMs \citep[][\textit{i.a.}]{hanna2023how, wang2023interpretability}
Therefore, it is entirely possible that our conclusions about the precise nature of computations carried out by LMs can be greatly strengthened when supplemented by the methods developed in these aforementioned works. 

\paragraph{Single Language} Finally, this work only focuses on property inheritance problems stated in the English language. This does little to contribute towards diversity in NLP research.

\section{Acknowledgments}
We thank Tom McCoy and Andrew Lampinen for providing comments on a previous version of the draft. (KM)$^2$ acknowledge funding from NSF Grant 2104995 awarded to Kyle Mahowald. We thank UT-Austin Linguistics for providing the hardware to run experiments.

\bibliography{anthology,custom}

\appendix

\section{Dataset and implementation details}
\label{sec:implementation}
Our experiments use the stimuli from \comps{}, released with an MIT License by \citet{misra-etal-2023-comps}, but with a modification that involves changing of the nonce words to obey the constraint that the in-context examples all have different nonce word pairs. To this end, we use the following nonce-words:
\begin{itemize}
    \item \textbf{In-context examples:} \textit{wug, dax, fep, zek, blick, toma, kiki, glorp, bova, zup, tufa, flib} (counter-balanced)
    \item \textbf{Test examples:} \textit{gek, wif} (counter-balanced)
\end{itemize}

\subsection{Methodological details}
Following \comps{}, as well as the precedent set by a number of previous minimal pair analyses \citep{linzen-etal-2016-assessing, gulordava-etal-2018-colorless, futrell-etal-2019-neural, wilcox-etal-2019-hierarchical, warstadt-etal-2020-blimp-benchmark, hu-etal-2020-systematic}, we use a forced choice task to evaluate our LM subjects. 
Like in \comps{}, we compare the log-probability of the property phrase (here, \textit{has a flat tail}) given the choice of left contexts (which indicate whether the right vs. the wrong concept has the property). For example, we measure:

\small{
    \begin{align*}
        \log P_{\theta}(\textit{has a flat tail} \mid & \textit{ a \textcolor{blu}{\textbf{gek}} is a beaver. a \textcolor{yell}{\textbf{wif}} is a} \\ 
        & \textit{ gorilla. therefore, a \textbf{\textcolor{blu}{gek}/\textcolor{yell}{wif}}}),
    \end{align*}
}

\normalsize
\noindent
and for \compsqa{}, we compare the relative probabilities of the two novel concepts given a fixed left prefix which contains a question about the property. For example, we measure:

\small
\begin{align*}
    \log P_{\theta}(\textit{\textbf{\textcolor{blu}{gek}/\textcolor{yell}{wif}}} \mid & \textit{ a \textcolor{blu}{\textbf{gek}} is a beaver. a \textcolor{yell}{\textbf{wif}} is a gorilla.} \\ 
    & \textit{ \textbf{Question:} Which one of them has a} \\
    & \textit{ flat tail? \textbf{Answer:}})
\end{align*}

\normalsize
\noindent
In both cases above, \textcolor{blu}{\textit{\textbf{gek}}} is the concept that should inherit the property. While these examples show the zero-shot case, cases with in-context examples and instructions simply add more context to the prefix, therefore the surface form of the output space remains the same regardless of the number of in-context examples or the presence of instructions.
Similar to the above measures, we measure the base property knowledge of LMs by comparing the conditional log-probabilities of the property phrases given the real, known concepts in question. That is, we measure:

\begin{align*}
    \log P_{\theta}(\textit{has a flat tail} \mid \textit{\textbf{beaver}/gorilla})
\end{align*}

The accuracy here, again, is the proportion of time the correct concept was more likely to inherit the property than the incorrect one. We measure this accuracy for all our test items zero-shot, and use that as the reference base property knowledge in order to contextualize our main results.

Log-probabilities for all models were accessed using \texttt{minicons} \citep{misra2022minicons},\footnote{\url{https://github.com/kanishkamisra/minicons}} a library that wraps around \texttt{transformers} \citep{wolf-etal-2020-transformers} by huggingface, and is written in \texttt{pytorch}. For our experiments with Llama-13B, we quantize the model to 4-bits in order to fit it onto a single GPU. All experiments were run on a cluster with 4 NVIDIA A40 GPUs, though each individual experiment on a model was computed on a single A40 GPU.


\subsection{Model Metadata}
\label{sec:model-metadata}

\begin{table*}[!t]
\centering
\begin{tabular}{@{}lrrr@{}}
\toprule
\textbf{Model} & \textbf{Params} & \textbf{Pre-training Tokens} & \textbf{Vocab size} \\ \midrule
GPT-2 XL \citep{radford2019language} & 1.5B & 8B & 50,257 \\
OPT-6.7b \citep{zhang2022opt} & 6.7B & 180B & 50,272 \\
Llama-2-7B \citep{touvron2023llama} & 7B & 1.8T & 32,000 \\
Mistral-7B \citep{jiang2023mistral} & 7B & ? & 32,000 \\ 
OLMo-7B \citep{groeneveld2024olmo} & 7B & 2.46T & 50,304 \\ 
Llama-3-8B \citep{dubey2024llama} & 8B & 15T & 128,000 \\ 
Gemma-2-9B \citep{team2024gemma} & 9B & 8T & 256,000 \\
Llama-2-13B \citep{touvron2023llama} & 13B & 1.8T & 32,000 \\
\bottomrule
\end{tabular}%
\caption{Overview of the non-instruction tuned LMs used in this work. `?' indicates that the given value was not made available in the LM's release. In addition to these, we also report results on instruction-tuned versions of \textbf{OLMo-7B}, \textbf{Mistral-7B}, \textbf{Llama-3-8B}, and \textbf{Gemma-2-9B}.} 
\label{tab:modelmeta}
\end{table*}

\Cref{tab:modelmeta} shows the LMs used in this work, along with their total parameters, tokens encountered during training, and vocabulary size.

\section{Instructions}
\label{sec:instructions}

Tables \ref{tab:minimal}, \ref{tab:aliens}, \ref{tab:inst1}, \ref{tab:inst2} show our instruction templates.

\input{instructiontables/minimal}
\input{instructiontables/aliens}
\input{instructiontables/inst1}
\input{instructiontables/inst2}

\section{Fine-grained results}
\label{sec:finegrained}
\begin{figure*}
    \centering
    \includegraphics[width=\textwidth]{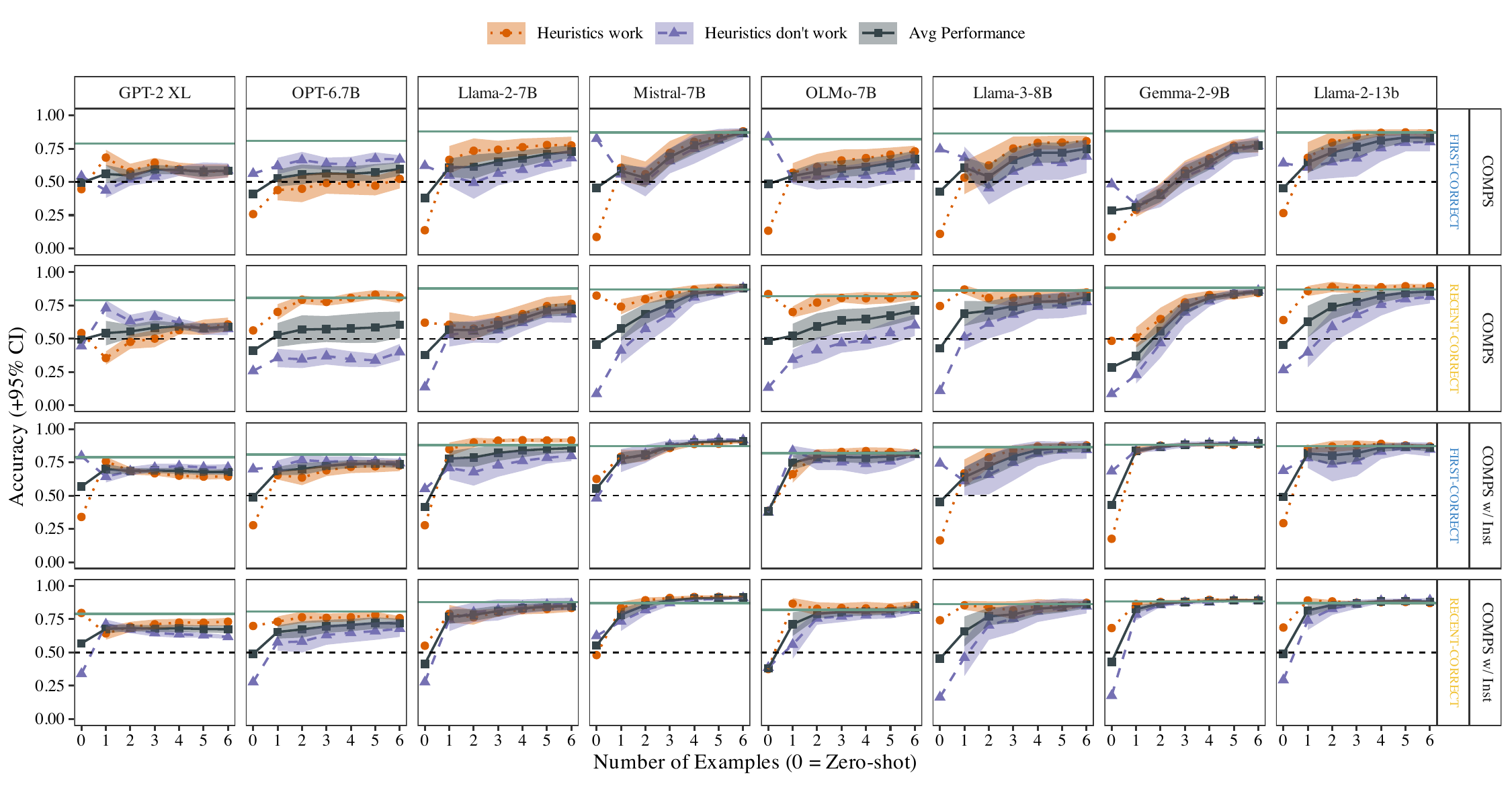}
    \caption{Fine-grained results for non-instruct tuned LMs on \comps{} as a function of the number of in-context examples (with and without instructions).}
    \label{fig:comps}
\end{figure*}

\begin{figure*}
    \centering
    \includegraphics[width=\textwidth]{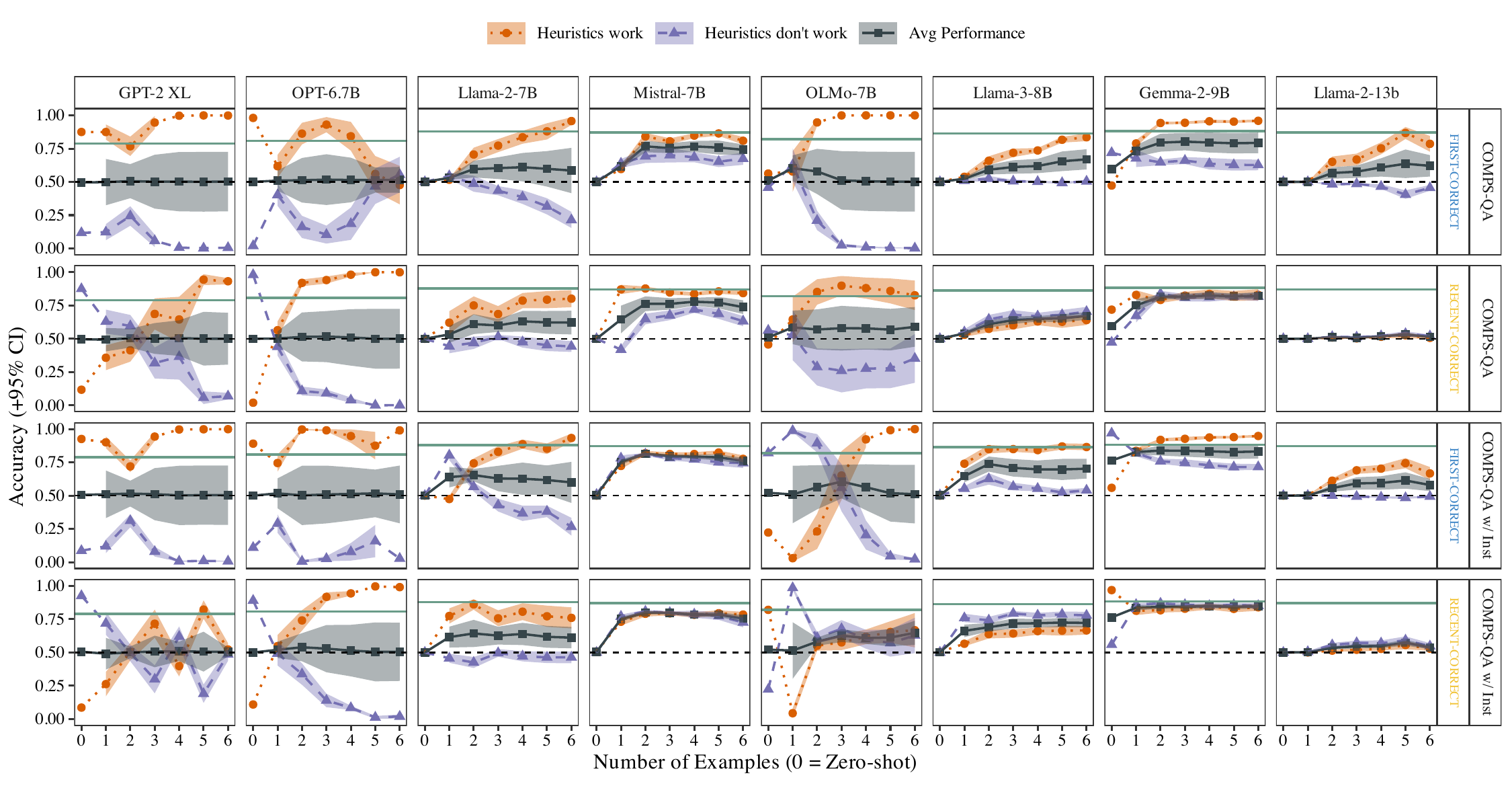}
    \caption{Fine-grained results for non-instruct tuned LMs on \compsqa{} as a function of the number of in-context examples (with and without instructions).}
    \label{fig:compsqa}
\end{figure*}

\begin{figure*}
    \centering
    \includegraphics[width=\textwidth]{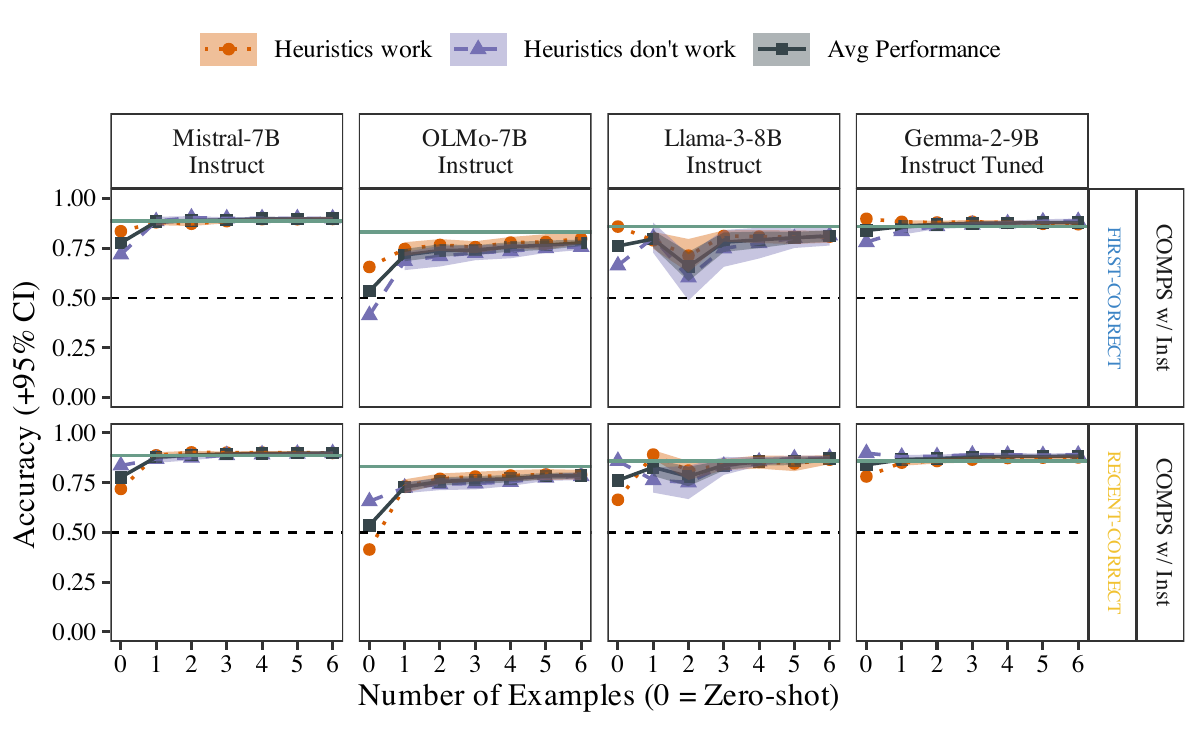}
    \caption{Fine-grained results for instruct tuned LMs on \comps{} as a function of the number of in-context examples (with and without instructions).}
    \label{fig:comps-instruct-lms}
\end{figure*}

\begin{figure*}
    \centering
    \includegraphics[width=\textwidth]{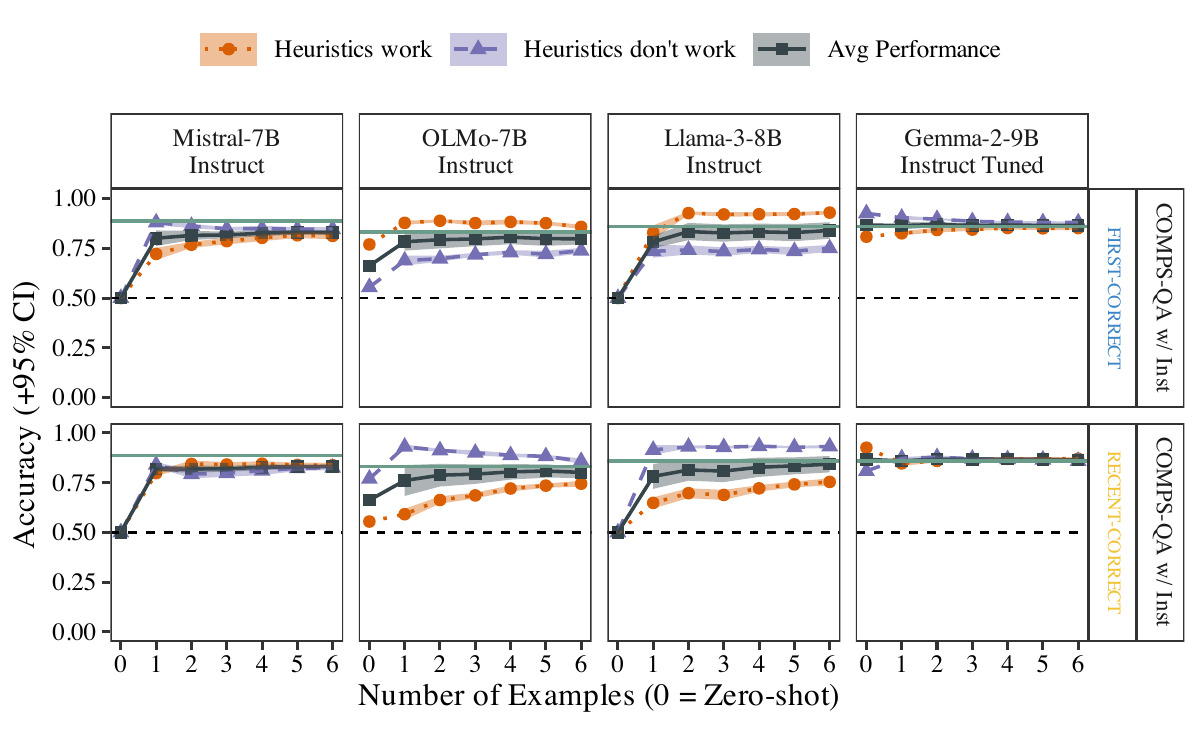}
    \caption{Fine-grained results for instruct tuned LMs on \compsqa{} as a function of the number of in-context examples (with and without instructions).}
    \label{fig:compsqa-instruct-lms}
\end{figure*}

While \Cref{fig:main-results} shows results aggregated over both types of heuristics that we have used in this work, we additionally display finer-grained results in this section, broken down by heuristic type.
Again, in each of these plots, the extent to which a model relies on heuristic is indicated by the gap between the \textcolor{orang}{dotted ($\newmoon$)} and the \textcolor{purpl}{dashed ($\blacktriangle$)} lines. This is now separately shown for each of our heuristics.
\Cref{fig:comps} shows results for non-instruct tuned LMs on \comps{} with and without instructions for both the heuristics, and similarly \Cref{fig:compsqa} shows results for those LMs on \compsqa{} with and without instructions for both the heuristics. \Cref{fig:comps-instruct-lms} and \Cref{fig:compsqa-instruct-lms} show analogous results for Instruction tuned LMs. In all these plots, the solid green line denotes the base performance of LMs.


\end{document}

%% file: instructiontables/minimal.tex
\begin{table*}[!ht]
\centering
\begin{tabular}{@{}lp{5in}@{}}
\toprule
\textbf{\comps{} version} & \textbf{Instruction Template} \\ \midrule
\comps{} & Given a pair of statements that introduce novel entities as types of real world animals, \textbf{write a true statement about the properties of the novel entities:}\newline\newline\texttt{\{examples\}} \textcolor{red}{(omitted in zero-shot)}\newline\texttt{\{test-stimulus\}} \\ \midrule
\compsqa{} & Given a pair of statements that introduce novel entities as types of real world animals, \textbf{answer the question that follows}:\newline\newline\texttt{\{examples\}} \textcolor{red}{(omitted in zero-shot)}\newline\texttt{\{test-stimulus\}} \\ \bottomrule
\end{tabular}%
\caption{Instructions for \comps{} and \compsqa{} with instruction type: ``\texttt{minimal}''}
\label{tab:minimal}
\end{table*}

%% file: instructiontables/aliens.tex
\begin{table*}[!ht]
\centering
\begin{tabular}{@{}lp{5in}@{}}
\toprule
\textbf{\comps{} version} & \textbf{Instruction Template} \\ \midrule
\comps{} & Some aliens have come to earth, and it turns out they have their own language for talking about our animals here on Earth. Your job is to help the aliens learn about our Earthling animals \textbf{by giving them some information about the animals.}\newline\newline Let's get started:\newline\texttt{\{examples\} }\textcolor{red}{(omitted in zero-shot)}\newline\texttt{\{test-stimulus\}} \\ \midrule
\compsqa{} & Some aliens have come to earth, and it turns out they have their own language for talking about our animals here on Earth. Your job is to help the aliens learn about our Earthling animals \textbf{by answering some questions about them.}\newline\newline Let's get started:\newline\texttt{\{examples\} }\textcolor{red}{(omitted in zero-shot)}\newline\texttt{\{test-stimulus\}} \\ \bottomrule
\end{tabular}%
\caption{Instructions for \comps{} and \compsqa{} with instruction type: ``\texttt{aliens}''}
\label{tab:aliens}
\end{table*}

%% file: instructiontables/inst1.tex
\begin{table*}[!ht]
\centering
\resizebox{\textwidth}{!}{%
\begin{tabular}{@{}llp{5in}@{}}
\toprule
\textbf{\comps{} version} & \textbf{No. of shots} & \textbf{Instruction Template} \\ \midrule
\multirow{2}{*}{\comps{}} & Zero-shot & It is important to know and reason about the properties of entities in the world. The following is a pair of premise statements that introduce novel entities as new types of real world animals. Your task is to make a conclusion about the properties of one of the entities by reasoning over the premise statements.\newline\newline \texttt{\{test\_stimulus\}} \\ \cmidrule(l){2-3}
 & Few-shot & It is important to know and reason about the properties of entities in the world. The following example(s) show a pair of premise statements that introduce novel entities as new types of real world animals, followed by another statement that attributes a property to one of the entities introduced in the premise statements.\newline\newline~Examples:\newline\texttt{\{examples\}}\newline\newline~Here is another pair of premise statements. Your task is to make a conclusion about the properties of one of the entities by reasoning over the premise statements.\newline\texttt{\{test\_stimulus\}} \\ \midrule
\multirow{2}{*}{\compsqa{}} & Zero-shot & It is important to know and reason about the properties of entities in the world. The following is a pair of premise statements that introduce novel entities as new types of real world objects. The statements are followed by a question that asks which novel entity in the premise can a specific property can be attributed to. Answer the question by reasoning over the premise statements.\newline\newline\texttt{\{test\_stimulus\}} \\ \cmidrule(l){2-3}
 & Few-shot & It is important to know and reason about the properties of entities in the world. The following example(s) show a pair of premise statements that introduce novel entities as new types of real world animals. The statements are followed by a question that asks which novel entity in the premise can a specific property can be attributed to, and the answer to the question, obtained by reasoning over the premise statements.\newline\newline~Examples:\newline\texttt{\{examples\}}\newline\newline~Here is another pair of premise statements. Answer the question that follows.\newline\texttt{\{test\_stimulus\}} \\ \bottomrule
\end{tabular}%
}
\caption{Instructions for \comps{} and \compsqa{} with instruction type: ``\texttt{Detailed-1}''}
\label{tab:inst1}
\end{table*}

%% file: instructiontables/inst2.tex
\begin{table*}[!ht]
\centering
\resizebox{\textwidth}{!}{%
\begin{tabular}{@{}llp{5in}@{}}
\toprule
\textbf{\comps{} version} & \textbf{No. of shots} & \textbf{Instruction Template} \\ \midrule
\multirow{2}{*}{\comps{}} & Zero-shot & It is important to know and reason about the properties of entities in the world. The following is a pair of premise statements that introduce novel entities as new types of real world animals. \textbf{Your task is to write a true statement about the properties of the novel entities.}\newline\newline \texttt{\{test\_stimulus\}} \\ \cmidrule(l){2-3}
 & Few-shot & It is important to know and reason about the properties of entities in the world. The following example(s) show a pair of premise statements that introduce novel entities as new types of real world animals, followed by another statement that attributes a property to one of the entities introduced in the premise statements.\newline\newline~Examples:\newline\texttt{\{examples\}}\newline\newline~Here is another pair of premise statements. \textbf{Your task is to write a true statement about the properties of the novel entities.}\newline\texttt{\{test\_stimulus\}} \\ \midrule
\multirow{2}{*}{\compsqa{}} & Zero-shot & It is important to know and reason about the properties of entities in the world. The following is a pair of premise statements that introduce novel entities as new types of real world animals. The statements are followed by a question that asks \textbf{which of the introduced entities} a specific property can be attributed to. Answer the question by reasoning over the premise statements.\newline\newline\texttt{\{test\_stimulus\}} \\ \cmidrule(l){2-3}
 & Few-shot & It is important to know and reason about the properties of entities in the world. The following example(s) show a pair of premise statements that introduce novel entities as new types of real world animals. The statements are followed by a question that asks \textbf{which of the introduced entities} a specific property can be attributed to, and the answer to the question, obtained by reasoning over the premise statements.\newline\newline~Examples:\newline\texttt{\{examples\}}\newline\newline~Here is another pair of premise statements. Answer the question that follows.\newline\texttt{\{test\_stimulus\}} \\ \bottomrule
\end{tabular}%
}
\caption{Instructions for \comps{} and \compsqa{} with instruction type: ``\texttt{Detailed-2}''}
\label{tab:inst2}
\end{table*}